\documentclass[5p]{elsarticle}
\usepackage[utf8]{inputenc}

\usepackage{lipsum}
\usepackage{graphicx}
\usepackage{color}
\usepackage{enumitem}
\usepackage{amssymb,amsmath,bm}
\usepackage{textcomp}
\usepackage{multirow}
\usepackage[hidelinks]{hyperref}
\usepackage{doi}
\usepackage{subcaption}
\usepackage{array}
\usepackage{float}
\usepackage{eurosym}
\usepackage{booktabs}
\usepackage{stfloats}
\usepackage{verbatim}

\begin{document}

\begin{frontmatter}
\title{Understanding electricity consumption behaviour through Inverse Reinforcement Learning}

\author[1,3,4]{E. Cofler\corref{cor1}} 
\ead{enrico.cofler@polimi.it}
\author[2,3,4]{C. Rodriguez-Pardo}
\author[1,3,4]{M. Giuliani}
\author[1,3,4]{A. Castelletti}
\author[2,3,4]{M. Tavoni}
\cortext[cor1]{Corresponding author. Present address: Dipartimento di Elettronica, Informazione e Bioingegneria, Politecnico di Milano, Via Ponzio 34/5, 20133 Milano, Italy}
\address[1]{Dipartimento di Elettronica, Informazione e Bioingegneria, Politecnico di Milano, Via Ponzio 34/5, 20133 Milano, Italy}
\address[2]{Dipartimento di Ingegneria Gestionale, Politecnico di Milano, Via Raffaele Lambruschini 4B, 20156 Milano, Italy}
\address[3]{RFF-CMCC European Institute on Economics and the Environment, Via Solari, 11, 20144 Milano, Italy}
\address[4]{Centro Euro-Mediterraneo sui Cambiamenti Climatici, Via Solari, 11, 20144 Milano, Italy}

\begin{abstract}
Understanding how households consume electricity in response to socioeconomic and climatic drivers is important for decision-makers designing energy policies in a changing climate and under geopolitical tensions. Consumers respond differently to thermal stress depending on income, consumption habits and the surrounding built environment, a nonlinear behaviour that most approaches oversimplify.
In this study, households are treated as agents interacting with complex environments, and Inverse Reinforcement Learning is used to represent their consumption behaviour as model-implied reward functions. Specifically, we observe how these reward functions change when households undergo socioeconomic and climatic shocks.
The framework is tested on different clusters of electricity consumption profiles in Italy. Clusters' reward functions are retrieved and used to understand how cooling behaviour changes from summer 2021 to summer 2022 and 2023, before, during and after the energy crisis and a heatwave. We find that these shocks reshaped cooling behaviour heterogeneously across consumer groups, in directions conditioned by their prior habits and built environment. Across the 2021–2023 summers, we identify a spectrum of responses: transient adjustments that receded as the shocks eased, durable shifts persisting into 2023, and consumers exhibiting negligible change. At the sub-daily scale, groups comparable in socioeconomic and environmental context but differing in their daily timing of consumption responded distinctly, identifying time-of-use as a separate dimension of behavioural heterogeneity. Energy policies and demand-response schemes should therefore account not only for who consumers are and where they live, but for when they consume and whether their response to a shock persists.
\end{abstract}


\end{frontmatter}

\section{Introduction}\label{sec:Introduction}
Residential building stock energy demand is a major component of global energy demand, accounting for 18\% of global electricity consumption. For buildings, electricity is also the fastest-growing energy carrier, with a rate of 161\% over the last 30 years \cite{intergovernmentalpanelonclimatechangeipccBuildings2023}. Furthermore, household electricity consumption has become a national concern. The energy crisis following the Russian invasion of Ukraine has had major repercussions for energy affordability, leading all governments to take unprecedented and costly steps to protect consumers from price spikes. At the same time, climate change has shifted the patterns of consumption, with increased cooling needs in hotter summers. Cooling energy needs have important ramifications for the electricity grid and for power generation, but also for household wellbeing.
Uncovering and forecasting the behaviour of households, especially as electricity consumers, is therefore a fundamental task to understand the impact of their energy demand on society, GHG emissions and on the grid. Households' social and technical heterogeneity is reflected in how they consume or avoid consuming energy \cite{congUnveilingHiddenEnergy2022}, but disentangling the behavioural signal from consumption data is a complex task, usually approached by oversimplifying it.
For example, to help the design of energy policies that take into account the stability of the energy grid, it is necessary to model the behaviour of users at short time-scales, such as sub-daily temporal resolution. However, current approaches make use of limiting modeling assumptions to tackle behaviour identification \cite{andersenPriceResponseResidential2026}, and struggle to take into account the many factors that drive the heterogeneity in consumption behaviours, such as activity schedules both inside and outside the house, socioeconomic status, energy prices, geographical location, and outdoor weather conditions. \\
One of the reasons for the lack of understanding of consumers' behaviours is the limited availability of data at a granular level. In recent years, a great amount of new electricity consumption data coming from smart meters has become accessible for research \cite{ushakovaBigDataRescue2020}\cite{vitielloSmartMeteringRollOut2022}. Big smart meter datasets provide an unprecedented level of detail that could enable disaggregated, highly precise modelling, also adequately representing the stochasticity present in empirical data at sub-daily resolution \cite{happleReviewOccupantBehavior2018}\cite{huebnerShapeWarmthTemperature2015}.
Traditional and explainable modelling approaches may, however, not fully capture the necessary complexity of the problem. White box models focus on stylized representations of the consumers to create archetypes that favour the explainability of results \cite{edelenboschTranslatingObservedHousehold2022}; on the other hand, more advanced optimisation and Machine Learning methods usually consider specific subpopulations, with limited numbers of consumers \cite{motlaghAnalysisHouseholdElectricity2015}\cite{uchidaAggregatedSmartMeter2025}. This constrains the generalizability of the results to whole countries or large-scale systems, and therefore such approaches struggle in providing insights that can be used for policy. 
Therefore, how to model household consumption on a large scale for broad consumer groups, while avoiding oversimplification and maintaining an intrinsic interpretability of the results, remains an open question. \\
An additional research gap emerges when considering the effects of the energy crisis specifically on space cooling energy demand and cooling behaviours.
As the literature has been focusing mostly on space heating \cite{einolanderDetectingChangesPricesensitivity2024}, and on average space cooling consumption responses fixed in time \cite{congUnveilingHiddenEnergy2022}\cite{harishImpactTemperatureElectricity2020},
the identification of cooling behaviours in terms of energy demand and temperature-response is limited, as well as the impact evaluation of shocks on cooling behaviours. Also, there is a limited understanding of the interaction with the surrounding environment, and dedicated modelling at sub-daily resolutions would be needed to appreciate the practice of space cooling in residential contexts in a more realistic framework.\\
In this paper, these research gaps are addressed using a novel approach based on Inverse Reinforcement Learning (IRL).
The set of instructions, according to which the agents act, is derived from an objective function that the agents try to maximize: the reward function.
Given the complexity in defining an appropriate reward function in many settings, the problem of Inverse Reinforcement Learning has been formulated so as to revert the initial formulation of RL and reverse-engineer a reward function from the available data \cite{ngAlgorithmsInverseReinforcement2000}. 
In this application, consumption profiles, along with a set of context attributes (such as urbanization level, temperature, bill cost), are used to compose a coherent RL environment. The aim is to retrieve the non-linear relationship between consumption and weather, considering also the socioeconomic and environmental context differences between consumers.\\ 
Within this framework, consumers, interpreted as agents, have information to take actions in the form of electricity consumption variations. They are separated into different clusters, which are then used as input to an IRL algorithm from which reward functions are retrieved.
\\Furthermore, the sub-daily resolution of the data is exploited to examine intraday behaviour: by sub-clustering on consumption timing, groups that are nearly indistinguishable in socioeconomic and environmental terms, yet differ in when they consume during the day, are isolated to understand whether intraday differences lead to different cooling behaviour responses during the energy crisis.\\
More specifically, this work employs an Adversarial Inverse Reinforcement Learning (AIRL) \cite{fuApplicationsReinforcementLearning2022} algorithm, which allows for a behaviour-oriented approach in the analysis of smart meter data, drawing on the existing literature on residential energy demand studies and applied IRL. This algorithm is known for its ability to produce complex reward functions that are disentangled from the environment dynamics in which they were learned \cite{sackmannModelingDriverBehavior2022}\cite{wangDecisionMakingAutonomous2021} and therefore transferable across changing dynamics and comparable: this is crucial for the observation and interpretation of rewards as model-implied representations of consumption behaviour.\\
The methodology is tested on Italian households' electricity smart meter data from May to July 2021, 2022, and 2023, before, during and after the energy crisis that has affected the European Union and led to a dramatic increase in wholesale electricity prices. 
We show that the clusters are well described by the reward functions: exposure to distinct environmental and socioeconomic contexts and the different energy consumption habits gives rise to a difference in how consumers react to thermal distress and specifically heat. Different types of behaviour shifts in consumers' rewards (persistent, transient, negligible) can be observed as the energy crisis unfolds and households experience higher electricity prices and higher temperatures. \\
The paper is organized as follow: Related Work (\ref{sec:RelatedWork}); Data (\ref{sec:Data}), with an overview on the variables considered in the model and data sources used for the Italian use case; Methods (\ref{sec:Methodology}), where the clustering and AIRL pipeline will be explained; Results (\ref{sec:Results}), presenting first the characteristics of the obtained clusters, and then analysing the selected clusters between 2021 and 2023; and Conclusions (\ref{sec:Conclusion}), summarising the main outcomes of the paper. 
\section{Related Work}\label{sec:RelatedWork}
In the literature, there is a good general understanding of how building energy consumption is related to the outside weather \cite{staffellGlobalModelHourly2023a}: models are able to reproduce the current data, although in many cases only at an aggregate level, and with limited explanations on the behavioural factors leading to the observed consumption. Also the effects of the interaction between prices, weather, and habits on residential energy consumption have been approached in recent publications \cite{andersenPriceResponseResidential2026}\cite{wangElectricityPriceHabits2021}\cite{burkhardtFieldExperimentalEvidence2023}. Results have generally not reached a consensus: energy prices, also from the experience of past energy crisis, do not seems to influence consumption in a consistent way, and underlying cultural and behavioural factors may be playing a major factor in determining this.\\
If, instead, the broader effects of socioeconomic shocks on consumption are considered, previous works from the literature report many results characterising the reaction of residential space heating energy demand. In particular, studies cover changes in price-sensitivity of households' electricity consumption \cite{einolanderDetectingChangesPricesensitivity2024}, demand adjustments due to the ending of contracts and price shocks   \cite{ahlvikHouseholdLevelResponsesEuropean}\cite{lunghiPowerPlayBalancing2026} and energy-saving behaviours implemented by households during extreme events \cite{pengDriversBarriersEnergysaving2026}.
Coming to cooling energy demand, most results do not focus on price and socioeconomic context as external shocks, but rather on short-term temperature response \cite{harishImpactTemperatureElectricity2020}, long-term projections and impacts on total consumption \cite{decianImpactAirConditioning2025}\cite{falchettaInequalitiesGlobalResidential2024}, and incentive-based demand response \cite{wangIncentiveBasedEmergency2023}\cite{burkhardtFieldExperimentalEvidence2023}.\\ The characterization of cooling behaviours in terms of energy demand and temperature response is therefore limited, as well as the impact evaluation of shocks on cooling behaviours: in other words, there is limited knowledge on how socioeconomic and climatic shocks affect how households' temperature response function changes, especially for space cooling. Some notable exceptions come from the field of energy poverty studies \cite{congUnveilingHiddenEnergy2022}\cite{kwonForgoneSummertimeComfort2023}\cite{huangInequalitiesCoolingHeating2023}, where, through the identification of cooling setpoints and cooling slopes, an estimate of cooling behaviours differences is obtained, also for different socioeconomic groups. To the authors' knowledge, only one such study focused on the energy crisis, using survey data to understand changes in households' (cooling) practices \cite{chatzikonstantinouHousingEnergyConsumption2022}. However, it does not quantify such behaviours in terms of energy demand, which is the focus of the current study.\\
Finally, the application of IRL to residential energy demand remains a relatively underexplored area in the literature. Recent work has begun to investigate imitation learning approaches for building energy management, although these studies have largely focused on operational decision-making \cite {deyInverseReinforcementLearning2023}\cite{liuExpertguidedImitationLearning2024}\cite{ImitationReinforcementLearning2026}. However, the technique has not yet been exploited to analyse behavioural differences in existing consumption data. AIRL has been applied prevalently in the field of autonomous driving and robotics, where it has been valued for recovering reward functions that remain valid when the environment dynamics change, rather than overfitting the specific conditions of the demonstrations \cite{sackmannModelingDriverBehavior2022}\cite{wangDecisionMakingAutonomous2021}. This is particularly important in those settings where the training environment could change, and it could be necessary to compare reward functions coming from different environments. This is also the case of households consuming electricity while at the same time navigating a changing climate and an energy system in transition and subject to severe shocks.\\
This paper makes three contributions to the literature. First, it introduces an Adversarial Inverse Reinforcement Learning framework to infer reward-implied behavioural response surfaces from large-scale household smart-meter data. Rather than interpreting observed electricity use only as an empirical temperature-consumption relationship, the proposed framework represents households as agents whose consumption variations are shaped by thermal, socioeconomic, environmental, and temporal context.\\
Second, the framework is applied to Italian residential electricity consumption during the summer periods of 2021, 2022, and 2023, a period spanning the pre-crisis, crisis, and post-peak phases of the European energy-price shock. This allows us to examine how cooling-related electricity consumption responses changed under concurrent thermal and socioeconomic stress. \\
Third, the paper investigates heterogeneity both across broad consumer clusters and within high-consumption groups at sub-daily scale. In doing so, it shows that differences in the timing of electricity use can reveal behavioural heterogeneity that is not captured by socioeconomic and environmental variables alone.

\section{Data}\label{sec:Data}
\begin{figure*}
    \centering
    \includegraphics[width=0.9\linewidth]{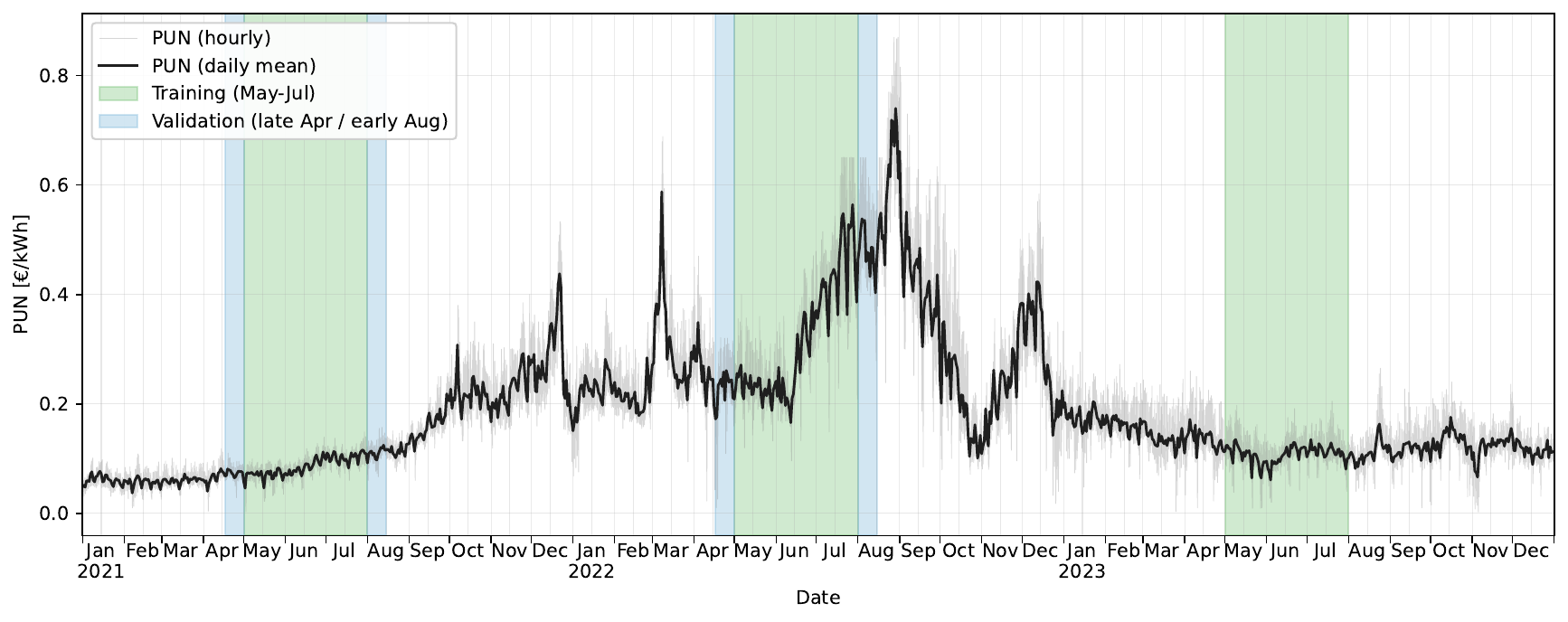}
    \caption{Daily (black) and hourly (grey) wholesale price values (PUN) between 2021 and 2023, periods covered by training data (green) and by validation data (blue).}
    \label{temporalOverview}
\end{figure*}

In this work, we use electricity metered data from smart meters for Italy.
Italy presents good coverage for second-generation smart meters in residential buildings, compared to most EU countries, which improves the data availability for this type of modelling exercise \cite{vitielloSmartMeteringRollOut2022}.  Also, it is characterized by a wide variety of climates and socio-economic conditions, and energy consumption habits \cite{besagniSociodemographicGeographicalDimensions2019}.
It goes from regions in the North where a large share of energy expenditure per capita is for gas and petroleum products, used for heating, to regions in the South where electricity and fossil fuels consumption shares are more balanced and space cooling is much more relevant \cite{campagnoloDistributionalConsequencesClimate2022}. 
Therefore, it is a good candidate to find, extract and analyse considerably different behaviours from large groups of consumers that are already metering their consumption at hourly resolution.  \\
The database covers the December 2019 - July 2023 period, but here the focus will be on May-July periods for years 2021, 2022 and 2023 to highlight differences in cooling behaviours across the three years in which wholesale electricity prices changed drastically due to the energy crisis. Both 2019 and 2020 were excluded from the analysis: the dataset contains few consumers for those years, and energy-use patterns were affected by the early phases of the COVID-19 pandemic in Italy \cite{ferrandoChangesEnergyUse2023a}\cite{bahmanyarImpactDifferentCOVID192020}. The last two weeks of April and the first two weeks of August will be used for validation on out-of-distribution (OOD) samples, with the exception of 2023, where dataset limitations made this infeasible. An overview of the training and validation periods, overimposed on the general trend of the energy crisis represented by the wholesale price values (National Single Price, PUN), is shown in Figure \ref{temporalOverview}. \\
Summer 2022 in Italy has also been one of the warmest in recent years \cite{TropicalIspra}, especially compared to summer 2021 and 2023, leading to a significant number of heat-related deaths \cite{ballesterHeatrelatedMortalityEurope2023}. The combination of socioeconomic and thermal stress for households will therefore be an aspect to keep in consideration in this work. \\
Preprocessing, as well as the characterization of the used variables, is necessary before considering the methodology.
First, the \textit{smart meter data consumption profiles}, which are at hourly resolution, are upscaled at 4-hour resolution to balance temporal granularity against the computational tractability of the problem. 
From the data, there is also information on the contracts associated with the users, in particular the time period in which they were active, the type of offer and the price components. This is used to reconstruct the total bill amount, for the energy component only, as seen by households, and most importantly the \textit{billed energy price} [\euro/kWh]. Coherently with the billing period, the frequency of this variable is monthly. \\
From the electricity consumption data aggregated to daily resolution, \textit{AC ownership} can also be estimated, following Chen et al. (2019) \cite{chenNewMethodUtilizing2019a}.
More specifically, it is estimated for 2021 data and kept constant during the following years: this simplification allows us to avoid the treatment of AC ownership estimations during an energy crisis (from 2022 onward), which would be flawed by energy conservation behaviours in budget-constrained households \cite{kwonForgoneSummertimeComfort2023}.\\
Finally, the location of users is known up to the postal code and municipality level. From this coarse geolocalization of consumers, it is possible to associate the following properties with the dataset:
\begin{itemize}
    \item \textit{income distribution} -
    obtained from the Italian National Statistics Institute (ISTAT), which discloses income-tax declarations on an annual basis, aggregated at municipal and zipcode level and across seven income brackets along with the mean income within each bracket. Income brackets are the following: < 10K\euro, 10-15K\euro, 15-26K\euro, 26-55K\euro, 55-75K\euro, 75-120K\euro, >120K\euro. Each consumer in the dataset is associated with the full income distribution data of its area;
    \item \textit{urbanization rate} - starting from the Global Human Layer Built Surface data at 100 x 100 m$^2$ resolution \cite{pesaresiGHSBUILTSR2023AGHS2023}, the distribution of urbanization values for each municipal and zipcode area is computed, ranging from 0 (no built-up surface area) to 10.000 m$^2$ (all built-up surface); 
    \item \textit{building vintage} - similarly for urbanization, building's age of construction is obtained from the Digital Building Stock Model \cite{martinezDBSMR2025EU2025} at 100 x 100 m$^2$ resolution, and aggregated at municipal and zipcode level. The dataset uses five epochs to categorize buildings: pre-1980, 1980–1990, 1990–2000, 2000–2010, 2010–2020, and therefore the aggregated values range in between 1950 (assumed as the average vintage value for the pre-1980 epoch) and 2015 (the average vintage for the newest epoch 2010-2020); 
    \item \textit{Green View Index (GVI)} - the data source for this variable already considers all areas in Italy directly at the municipal and zipcode level \cite{falchettaTrackingGreenSpace2025}. It is a street green space estimator reporting the percentage of street-level 360° imagery occupied by canopy;
    \item \textit{Universal Thermal Comfort Index (UTCI)} \cite{jianHighTemporalResolution2024} is a multivariate parameter taking into consideration the local weather and its impact on the human body, to compute a measure of thermal comfort valid across different climatic conditions. There are 10 UTCI thermal stress categories, the ones of interest for this study are: above +46, extreme heat stress; +38 to +46, very strong heat stress; +32 to +38, strong heat stress; +26 to +32, moderate heat stress; +9 to +26, no thermal stress. Originally at 0.1° x 0.1° spatial resolution and hourly resolution, we upscale it at the zipcode and municipal level, as done for the other variables considered in this list, and at 4-hour resolution, as done for the smart meter data.
\end{itemize}
Variables for which no time series data were available (urbanization rate, building vintage, GVI) are treated as constant over the three-year period under review. 
\section{Methodology}\label{sec:Methodology}
\begin{figure*}[t]
    \centering
    \includegraphics[width=\linewidth]{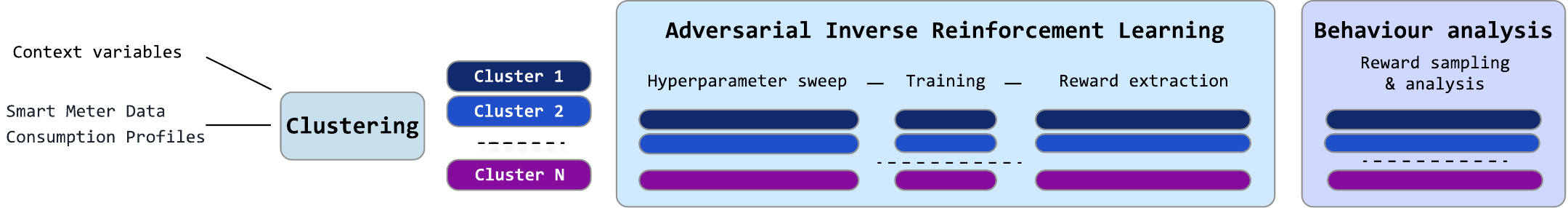}
    \caption{Full methodology pipeline, divided into (from left to right) Clustering on input data, AIRL training and hyperparameter optimization, and Behaviour analysis on the extracted rewards.}
    \label{methodGeneral}
\end{figure*}

The general methodology pipeline is shown in Figure \ref{methodGeneral}: the following subsections, Clustering, AIRL and Behaviour analysis, will explain in detail its subcomponents.
\subsection{Clustering}\label{sec:Clustering}
\begin{figure*}[t]
    \centering
    \includegraphics[width=0.87\linewidth]{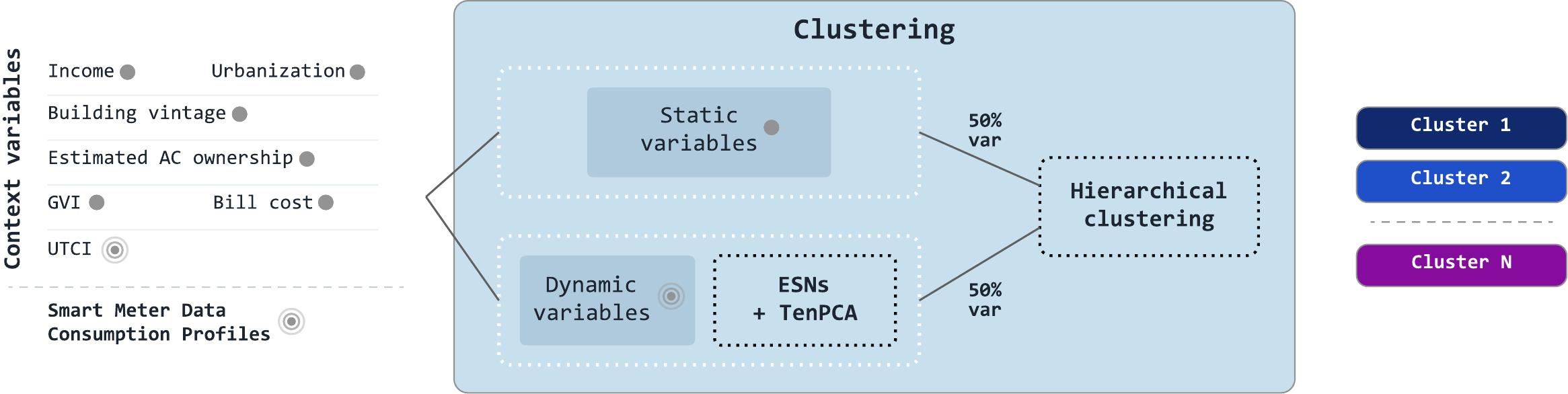}
    \caption{Clustering methodology, where the different pipelines for static/quasi-static and dynamic variables is shown. The output of this phase is the clusters, also shown in Figure \ref{methodGeneral}.}
\label{methodCluster}
\end{figure*}

The clustering pipeline, shown in Figure \ref{methodCluster}, is defined to capture different features of interest while maintaining a manageable set of clusters with a sufficient number of consumers and data points each.
In order to create a distance metric upon which clustering can be performed, variables are separated into static or quasi-static and time-dependent. In the first group, there are variables such as income, urbanization, building vintage, estimated AC ownership, GVI and bill cost, which are either constant or varying at monthly or yearly frequency. \\ Time-dependent variables consist of electricity consumption and UTCI, both at 4-hour resolution. They are treated so as to have a fixed-length representation that preserves temporal dynamics, while keeping computational cost limited. To do so, Echo State Networks (ESNs) \cite{bianchiReservoirComputingApproaches2021} are used: this approach is based on recurrent neural networks where the recurrent part is generated randomly and kept fixed, and subsequently used to learn the dynamic features of multivariate time series. The original dynamics is then mapped to a low-dimensional embedding to avoid overfitting and reduce the size of the problem for computational tractability. To do so, Tensor PCA \cite{koldaTensorDecompositionsApplications2009} is used, which generalises PCA \cite{PrincipalComponentAnalysis2002} from matrices to three-way data structures, simultaneously factorising the sample, time, and feature dimensions. Therefore, ESNs method works as an unsupervised nonlinear feature extractor for multivariate time series. In our case, the time series are the consumer's bivariate sequence (UTCI and consumption time series). The result of this extraction is then standardized, as done for the static/quasi-static variables.\\
The two groups of variables are then concatenated so that 50\% of the variance is attributed to each group, and pairwise Euclidean distances are computed.
Finally, the distance array obtained with this approach is subject to agglomerative hierarchical clustering \cite{wardHierarchicalGroupingOptimize1963} adopting Ward linkage \cite{virtanenSciPy10Fundamental2020}. The clusters obtained in this way become the input for the AIRL framework.

\subsection{Adversarial Inverse Reinforcement Learning}

\subsubsection{Model description}

The objective of IRL is to obtain a latent reward function $r(s, a)$ that explains the behaviour of an expert demonstrator, assuming that the expert actually performs optimally with respect to the given reward in a Markov Decision Process  $\mathcal{M} \setminus r = \langle \mathcal{S}, \mathcal{A}, \mathcal{T}, \gamma \rangle$, with states $s \in \mathcal{S}$, actions $a \in \mathcal{A}$, the transition function $T(s_{t+1} \mid s_t, a_t)$ and the discount factor $\gamma$. \\
AIRL \cite{fuLearningRobustRewards2018} frames this reward recovery in an adversarial game between a generator policy and a discriminator, and aims at disentangling the learned reward from the single environment in which it operates, to generalize it. A schematic representation of the model can be observed in Figure \ref{methodAIRL}. The generator policy is trained using Proximal Policy Optimization (PPO) \cite{schulmanProximalPolicyOptimization2017} so as to maximize the discriminator-assigned reward, which corresponds to maximizing the likelihood for the generator of being classified as an expert.\\
Furthermore, the reward is a neural network and acts as the AIRL discriminator, and it is composed by a base term $f_\phi(s, a)$ and a potential-based shaping term $\Phi_\psi(s)$:
$$r(s, a, s') = f_\phi(s, a) + \gamma \Phi_\psi(s') - \Phi_\psi(s)$$
The shaping term does not alter the optimal policy but adds capacity and stabilizes training; only $f_\phi$ is used at evaluation time. This decomposition is the core AIRL construction: it absorbs the policy- and dynamics- dependent component of the discriminator so that the extracted reward is consistent with observed trajectories rather than policy artefacts. This, however, does not imply a univocal identification of the reward. The problem is partially identified, as by construction multiple reward functions can explain the same observed behaviour. We therefore do not interpret $f_\phi$ as a direct psychological measure of household preferences, but as a compact, model-implied representation of consumption behaviour that is compatible with the observed trajectories.\\ 
Following the general workflow in Figure \ref{methodGeneral}, each cluster undergoes independent AIRL training, so as to extract its behaviour in the form of the reward function and highlight the differences between the different clusters. By clustering on all variables at once, we also assume that each cluster has an internally coherent environment and represents a single behaviour \cite{trauthLearningAdaptingBehavior2023}. This is a crucial assumption in this setup, as it highly simplifies the problem at hand by avoiding the need of multiple-intention modeling approaches \cite{likmetaDealingMultipleExperts2021a}.\\
Within each cluster, each consumer represents a trajectory $\tau^{(i)} = (s_0^{(i)}, a_0^{(i)}, s_1^{(i)}, \ldots, s_{T-1}^{(i)}, a_{T-1}^{(i)}, s_T^{(i)})$ of length $T$, where $T$ is the number of sub-daily intervals in the observation window and the states contain all the contextual information. Actions are defined as the first-order difference of the consumption time series $a_t = e_{t+1} - e_t$: in other words, consumption variation is the only action that each household can perform in this framework, while being affected by all variables contained in the state.
Three temporal variables representing the time signal are also included: time of the day, day of the year, and day type (working day or weekend/public holiday).
Furthermore, the agents are provided with a 24-hour rolling mean of UTCI and the four individual past consumption values as explicit memory features, so as to help in learning the temporal features of consumption behaviour.

\subsubsection{Training Methodology}
\begin{figure}[t]
    \centering
 \includegraphics[width=0.8\linewidth]{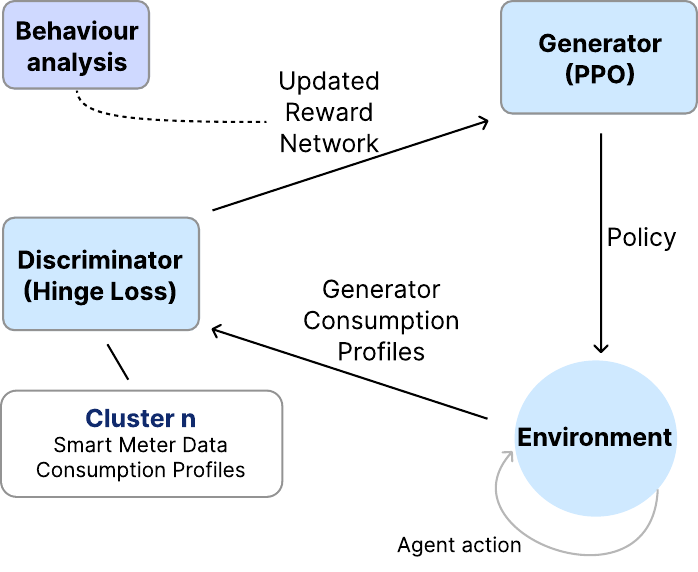}
    \caption{Schematic representation of the AIRL learning process in this study's implementation, followed by the reward extraction to perform the Behaviour analysis as anticipated in Figure \ref{methodGeneral}.}
\label{methodAIRL}
\end{figure}

Training alternates between generator and discriminator updates in an adversarial loop, as shown in Figure \ref{methodAIRL}.
Model quality is assessed periodically during training and at convergence by comparing the distribution of generated consumption trajectories against the held-out expert demonstrations. The following metrics are computed: distribution similarity between generated and expert consumption values $W_e$, distribution similarity between generator and expert actions $W_a$, total consumption error of the generator $\Delta E$, temporal correlation computed at 4-hour resolution $\rho_{4h}$ and temporal correlation at daily resolution $\rho_{\text{daily}}$ (by summing over the 6 intra-day timesteps).
Weighing these values, a final Combined Metric (CM) is defined so that it is to be minimized through the hyperparameters search:
\begin{multline*}
        CM = \alpha W_e + \beta \Delta E + \gamma W_a + \\
        + \eta(1 - \rho_{4h}) + \zeta(1 - \rho_{\text{daily}})
\end{multline*}
with coefficients $\alpha$, $\beta$, $\gamma$, $\eta$, $\zeta$ described more in detail in the Supplementary Materials. These metrics are used to validate the model's results and to perform an extensive hyperparameters selection, conducted using the Weights \& Biases sweep framework \cite{wandb} with Bayesian optimisation as the search strategy \cite{snoekPracticalBayesianOptimization2012}. 
\subsubsection{Implementation}
In this work, the AIRL approach is developed starting from the implementation of imitation-learning \cite{imitation-learning-gleave}, with a Gymnasium-compatible environment \cite{towersGymnasiumStandardInterface2025}.
All network parameters are optimized with AdamW \cite{loshchilovDecoupledWeightDecay2019}.

\subsubsection*{Generator}
The PPO architecture has a shared-input and dual-network actor-critic architecture. The observation vector is forwarded independently to two separate Multi-Layer Perceptrons: the policy network, which parametrizes the action distribution $\pi_\theta$, and the value network, which estimates the state-value function $V_\theta(s_t)$ \cite{schulmanProximalPolicyOptimization2017}. Both networks share the same structural design, each consisting of two hidden layers with a configurable activation function.
The action scale is also subject to annealing to prevent the generator from exploring the full action range from the beginning of the training: starting at 10\% of the full action scale, it is progressively amplified according to a trainable annealing parameter to reach the full action scale during training.

\subsubsection*{Discriminator}
In the current implementation of the discriminator's reward network, the following methodological choices are applied: spectral normalization \cite{miyatoSpectralNormalizationGenerative2018} to hidden layers to constrain the network's weight growth; batch normalization \cite{ioffeBatchNormalizationAccelerating2015} to stabilize the input distribution to each layer; orthogonal weight initialization \cite{engstromImplementationMattersDeep2020} to ensure a well-conditioned state at the beginning of the training.
Also, the dropout of a fraction $p$ of neurons to zero helps the generator to avoid overfitting on early-stage rewards produced by the discriminator \cite{isolaImagetoImageTranslationConditional2018}. \\
The discriminator loss has also been modified with respect to the original AIRL formulation: a hinge loss with label smoothing (targets 0.1/0.9) \cite{limGeometricGAN2017} is used.
The updated loss, unlike the binary cross-entropy loss originally used in AIRL, zeroes out the discriminator gradient once expert and generator logits are separated by a sufficient margin, preventing the discriminator from dominating the generator and preserving a meaningful policy gradient signal. Label smoothing softens the hard 0/1 targets, reducing overconfident discriminator outputs and improving training stability \cite{salimansImprovedTechniquesTraining2016}.
Gradient clipping \cite{pascanuDifficultyTrainingRecurrent2013} and a linear learning-rate warm-up \cite{goyalAccurateLargeMinibatch2018} are further applied to prevent the discriminator from dominating over the generator in the early stages of training, when the generator policy is still close to random.
To prevent the discriminator from memorizing superficial statistical artifacts of the expert data, Gaussian noise is injected into the  inputs at each forward pass, where the noise scale is itself a learnable parameter applied only to dynamic variables \cite{arjovskyPrincipledMethodsTraining2017}. \\
Finally, when static context variables risk overwhelming the dynamic consumption signal, an optional encoder multi-layer perceptron, which is optimized along with the discriminator, compresses them into a low-dimensional embedding. The reward network then receives the dynamic features concatenated with this embedding, enforcing a rebalance between behavioural dynamics and socioeconomic context.

\subsection{Behaviour analysis}
Once the reward networks $R_\theta(s, a)$ are trained for each cluster, their internal structures are interrogated directly to extract behavioural insights. To account for the stochastic nature of the learned behaviour and to calibrate the reward sampling to the empirical consumption mean, a softmax distribution over consumption values is constructed at temperature $\tau$:
$$\pi_\tau(e \mid s) \propto \exp\left(\frac{R_\theta(s[e], 0)}{\tau}\right)$$
The outer expectation over $s$ is approximated by Monte-Carlo sampling: a random sample $\{{s_i}\}_{i=1}^N$ is drawn from the expert observations, the per-state softmax expectation $\mathbb{E}_\tau[e \mid s_i]$ is evaluated at each, and $\tau$ is found so that $\frac{1}{N}\sum_{i=1}^N \mathbb{E}_\tau[e \mid s_i] = \bar{e}_\text{emp}$, with $\bar{e}_\text{emp}$ the empirical mean consumption.\\
The calibrated surface $\mathbb{E}_{\tau}[e \mid s]$ constitutes the primary output for the behaviour analysis. It is instrumental to visualize the learned reward structure and the overall optimal consumption states that the agents can choose depending on different values of thermal stress.\\
The second measure used to guide the analysis is simply the average 4-hour consumption for each UTCI value: this measure is particularly useful to understand how the reward function signal is reflected in the empirical consumption, or how they differ, and it makes no use of the AIRL pipeline.
\subsection{Validation on out-of-distribution samples}\label{sec:ValidationOOD}
Given the focus on cooling behaviours, the methodology previously presented has to be validated on periods immediately adjacent to the summer months used for training. In particular, two windows of two weeks before and after the training data periods have been considered, when data availability has made it possible. As already mentioned in the Data section (\ref{sec:Data}), OOD windows correspond to the last two weeks of April and the first two weeks of August, framing the May-June training period. The rationale behind the choice of two weeks is given by the tradeoff between collecting long-enough data time series to evaluate medium-term temporal correlations, while remaining close enough to the training window that the underlying consumer behaviour can be assumed unchanged. It should be noted that this assumption is most strongly tested in the August window, which is characterized by seasonal holidays and pronounced changes in activity intensity across many households in Italy \cite{istat_viaggi_vacanze}.\\
The aim of this test on OOD samples is to isolate the generalizability of the reward function from that of the policy. To this end, the reward network learned during training is frozen, and a new policy is fit on the OOD window under this fixed reward. The re-trained policy is then rolled out and evaluated against the OOD expert trajectories using the same Combined Metric adopted at training time. Because this procedure depends on the availability of contiguous data on both sides of the training window, the OOD validation could only be carried out for the years in which such adjacent data were available. \\ A complete presentation of these tests can be found in the Supplementary Materials. 
\begin{figure*}[!h]
    \centering
    \includegraphics[width=1\linewidth]{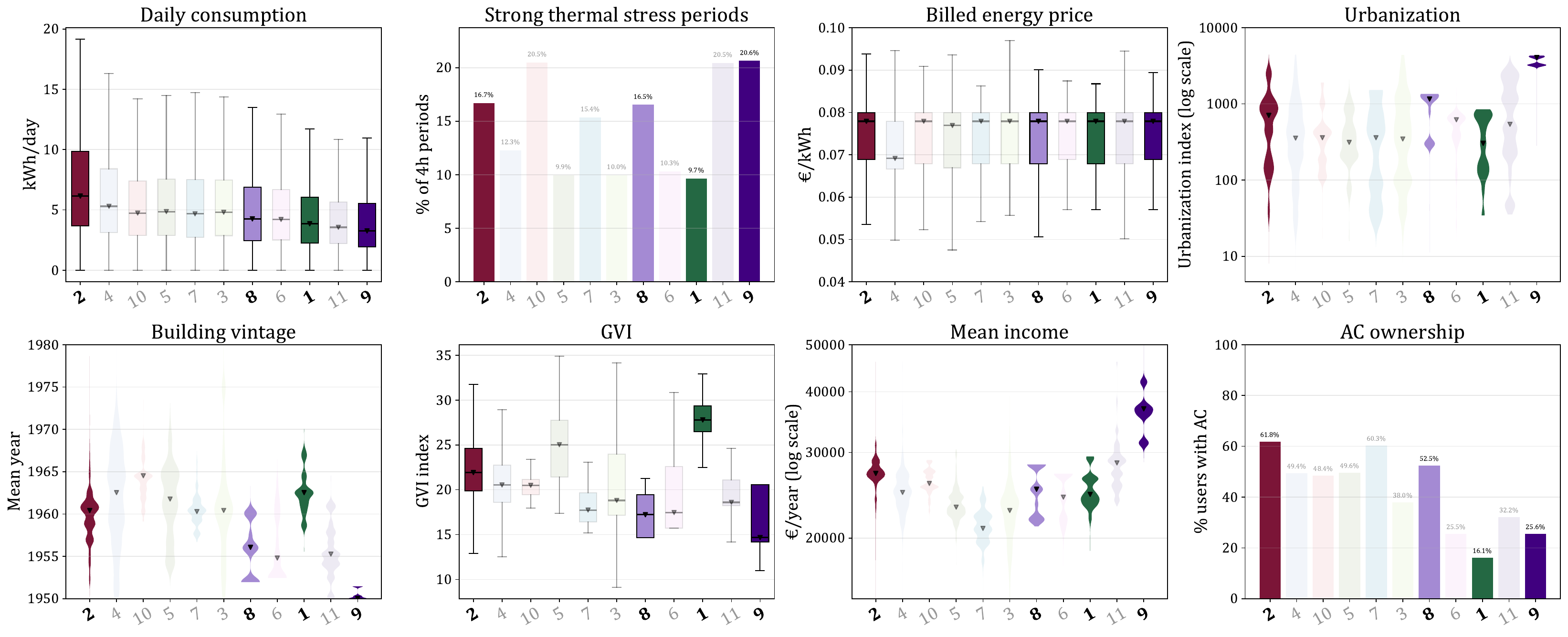}
    \caption{Comparison between clusters' characteristics in 2021, sorted from left to right for decreasing average daily consumption (top left). Clusters selected for further analysis are highlighted.  Top middle left: percentage of 4-hours period over 32° UTCI (strong heat stress region); top middle right: billed energy prices distribution [\euro/kWh]; top right: urbanization distribution in logarithmic scale [built surface $m^2$/(100$m$x100$m$)]; bottom left: buildings vintages distribution; bottom middle left: GVI distribution; bottom middle right: annual income distribution in logarithmic scale; bottom right: estimated percentage of consumers owning an AC. As mentioned in the text, the focus in the following analysis will be on cluster 2 (high-consumption), cluster 9 (urban intermediate-consumption, and cluster 1 (rural low-consumption).}
    \label{clusterCompareFig}
\end{figure*}

\section{Results}\label{sec:Results}
\subsection{Clustering}\label{sec:ResultsClustering}
We apply the clustering pipeline to May-July 2021, 2022, and 2023 data. The resulting 11 clusters range from 24,088 (cluster 11) to 3,098 (cluster 7) users each.
Clusters present broad discernible differences as shown in Figure \ref{clusterCompareFig}, where the context variables have been summarized for all clusters, ordered by average daily consumption.
Some clusters stand out for specific characteristics: for example, cluster 2 shows the highest average total consumption, as well as the highest estimated AC ownership, and also relatively high urbanization levels; cluster 9, on the other hand, shows the lowest consumption, matched with the highest urbanization and income levels of the areas it populates, indicating a very urban cluster; cluster 1 instead reports the lowest AC ownership, combined with very low consumption, as well as the highest GVI and the lowest thermal stress likely associated with its location in rural areas. Also, cluster 7 is found in the lowest income areas on average, while cluster 4 consumers see the lowest billed energy prices. Finally, other clusters such as cluster 8 and 6 report relatively average values for most of the metrics, including daily consumption, building vintage, AC ownership, and income. More information on clusters' spatial distribution can be found in the Supplementary Materials.\\
To limit the computational costs, we apply the AIRL pipeline only to a handful of clusters chosen to highlight a good share of the behavioural differences that can be observed through the analysis, with a particular focus on cooling: the \textbf{high-consumption} cluster (cluster 2), the \textbf{urban intermediate-consumption} cluster (cluster 8), \textbf{urban low-consumption} cluster (cluster 9) and the \textbf{rural low-consumption} cluster (cluster 1).

\begin{figure*}[!t]
    \centering
    \includegraphics[width=\linewidth]{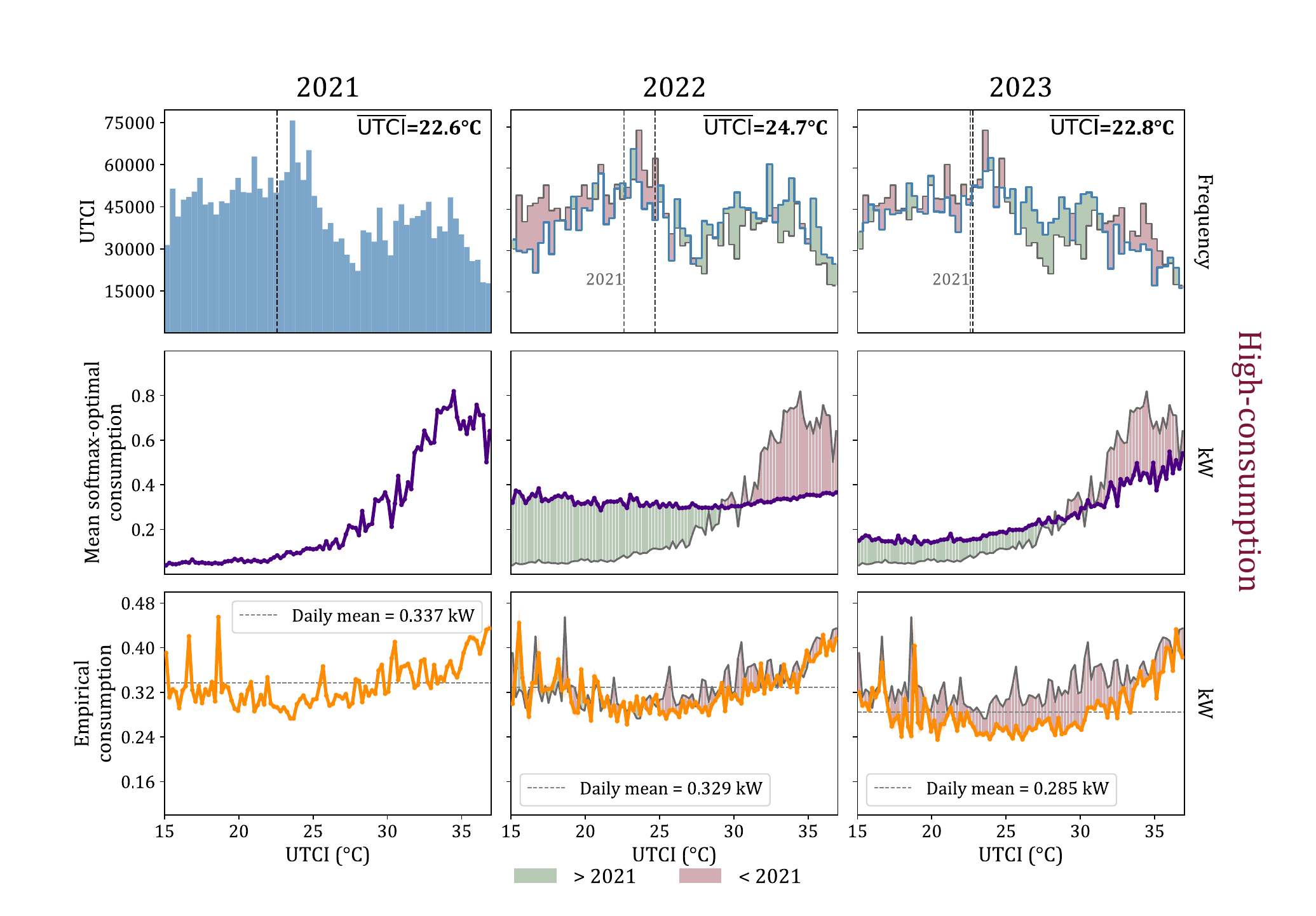}
    \caption{High-consumption cluster behaviour analysis, across May-June data for 2021 (left), 2022 (middle), 2023 (right). Top row reports the UTCI distribution experienced by the cluster's consumers; middle row shows the projection of the reward surface along consumption and UTCI, effectively showing the optimal consumption through different thermal stress levels according to the AIRL-retrieved reward; bottom row represents the empirical average 4-hours consumption vs UTCI (with light-coloured shading representing the standard error of mean consumption), computed directly from the observed data and therefore without using the AIRL pipeline. For each of these variables, 2022 and 2023 report the absolute trend and the variations with respect to 2021 values. Notice that for visualization purposes the empirical consumption scale is set to the range [0, 0.5] kW, while the softmax-optimal one to [0, 1].}
    \label{HighConsAnalysis}
\end{figure*}

\begin{figure*}[!t]
    \centering
    \includegraphics[width=\linewidth]{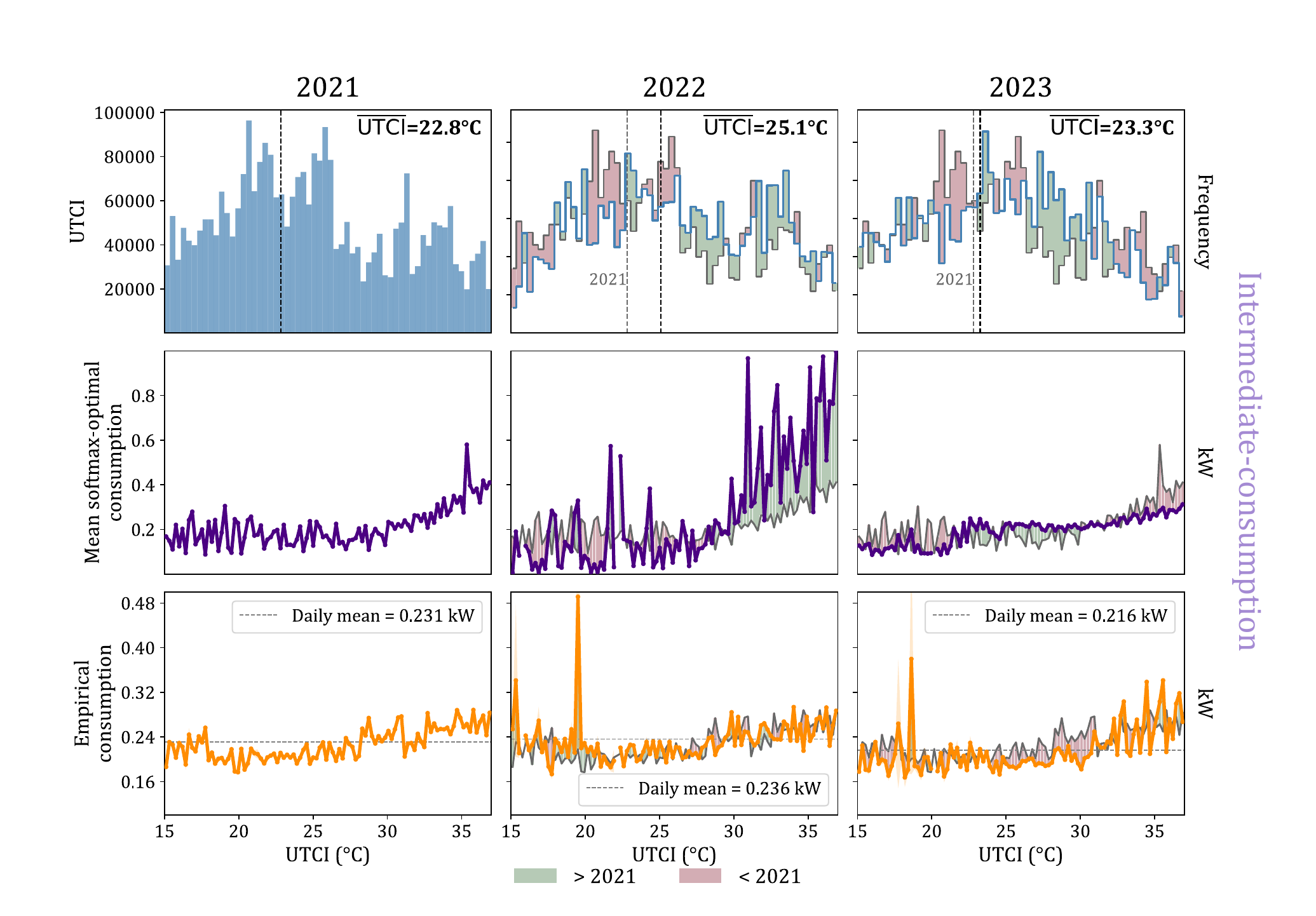}
    \caption{Urban intermediate-consumption cluster behaviour analysis, across May-June data for 2021 (left), 2022 (middle), 2023 (right). Legend description follows the one for Figure \ref{HighConsAnalysis}.}
    \label{UrbanIntermediateConsAnalysis}
\end{figure*}

\begin{figure*}[!t]
    \centering
    \includegraphics[width=\linewidth]{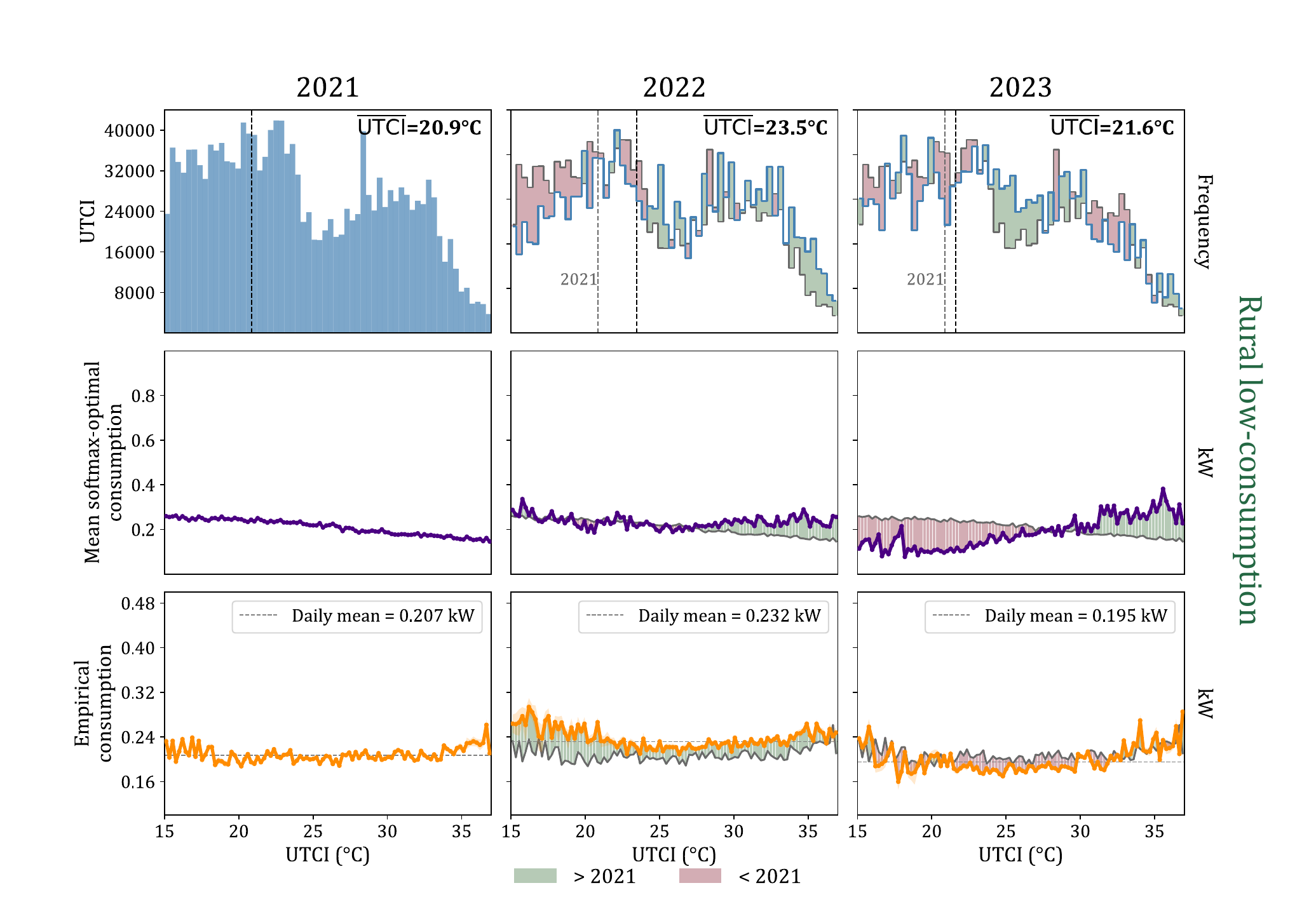}
    \caption{Rural low-consumption cluster behaviour analysis, across May-June data for 2021 (left), 2022 (middle), 2023 (right). Legend description follows the one for Figure \ref{HighConsAnalysis}.}
    \label{RuralLowConsAnalysis}
\end{figure*}

\begin{figure*}[!t]
    \centering
    \includegraphics[width=\linewidth]{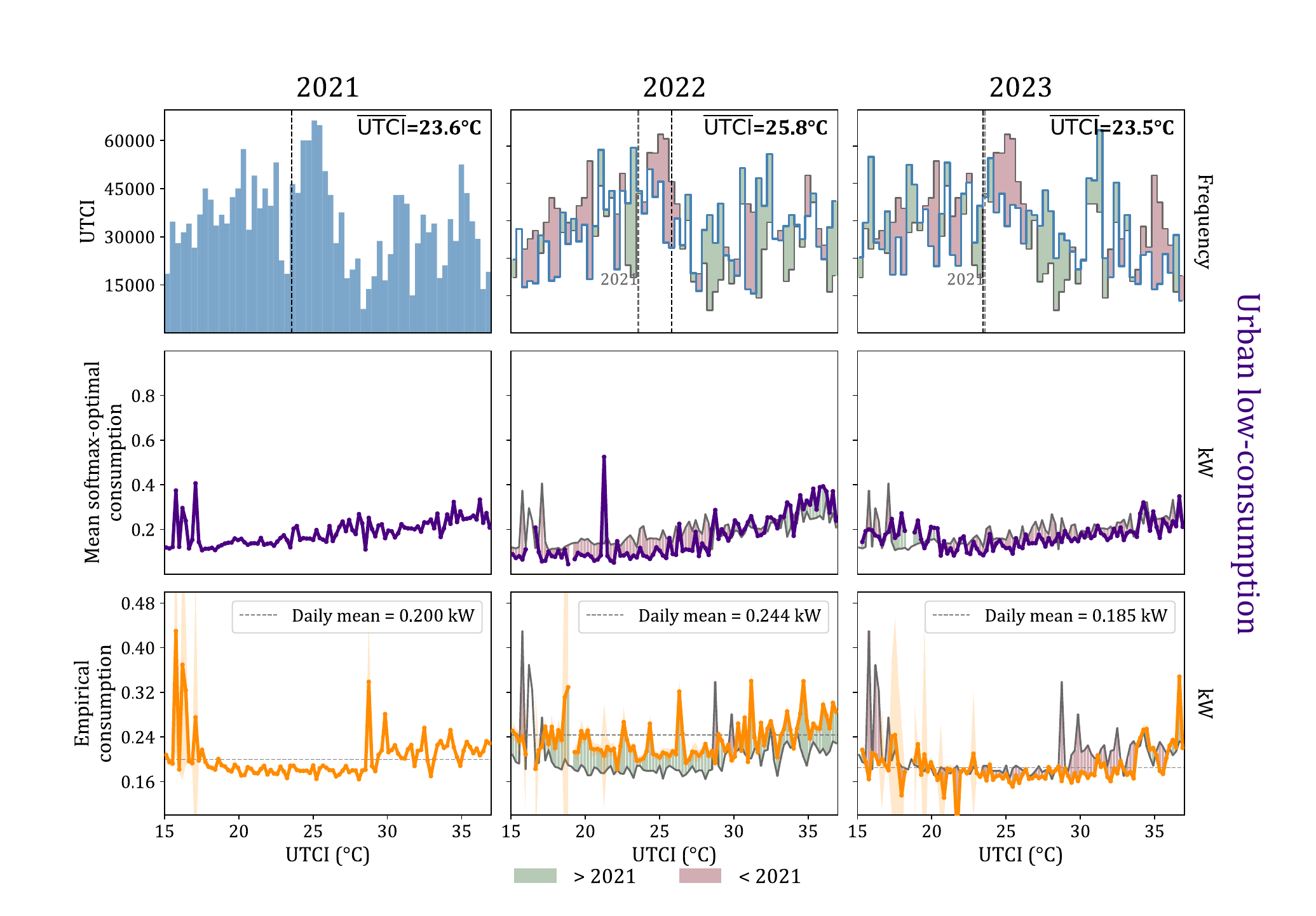}
    \caption{urban low-consumption cluster behaviour analysis, across May-June data for 2021 (left), 2022 (middle), 2023 (right). Legend description follows the one for Figure \ref{HighConsAnalysis}.}
    \label{UrbanLowConsAnalysis}
\end{figure*}

\begin{figure*}[!h]
    \centering
    \includegraphics[width=0.95\linewidth]{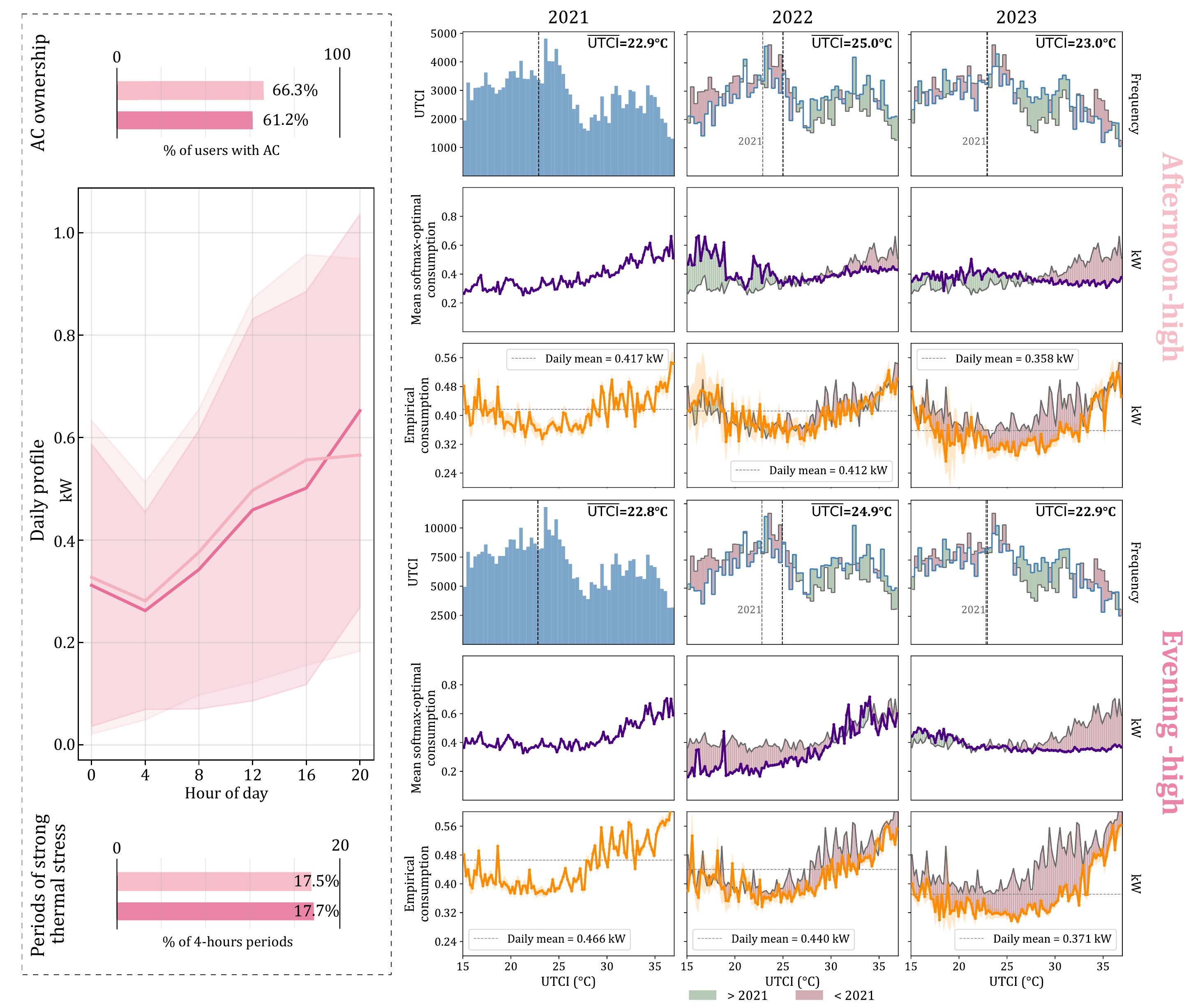}
    \caption{Analysis of high-consumption cluster's sub-clusters 4 (light pink) and 6 (dark pink). Top left: estimated percentage of consumers owning an AC; middle left: mean daily profiles; bottom left: percentage of periods of strong thermal stress experienced by the clusters; center and right top/bottom columns: sub-cluster 4/6 analysis for year 2021 and 2022, with legend description following the one for Figure \ref{HighConsAnalysis}, expect for the empirical consumption scale, which in this case it has been extended to considered the high consumption values observed by the two sub-clusters.}
    \label{intradailyDifferences}
\end{figure*} 

\subsection{Behaviour analysis}
As mentioned in the Methodology, the key variables to be analyzed in order to observe and understand behavioural changes in the considered clusters are two: the empirical average 4-hour electricity consumption for different values of thermal stress, and the corresponding optimal consumption levels computed from the reward function. These variables and their variations from 2021 to 2023, as well as the UTCI distribution, can be observed for the high-consumption cluster in Figure \ref{HighConsAnalysis}, for the urban intermediate-consumption one in Figure \ref{UrbanIntermediateConsAnalysis}, for the rural low-consumption one in Figure \ref{RuralLowConsAnalysis} and for the urban low-consumption one in Figure \ref{UrbanLowConsAnalysis}.
In general, as the energy crisis was reaching its peak (see Figure \ref{temporalOverview}) in 2022, all four clusters experienced higher thermal stress: indeed, as mentioned in the Introduction (\ref{sec:Introduction}), that year has been characterized by particularly high temperatures in Italy \cite{TropicalIspra}. However, the reaction to this combination of high UTCI and socioeconomic distress has not been the same.
Starting from the high-consumption cluster (estimated AC ownership 61.8\%), in 2021 both the reward's optimal consumption and the empirical one are strongly reacting to temperature, as it is expected for a cluster with such high AC penetration. Then, in 2022, compared to 2021, the empirical consumption slightly decreases, and this decrease is much more marked and differentiated for the optimal consumption levels: at low thermal stress, an increase in consumption is observed, while above 30° UTCI the opposite is true. Furthermore, this substantial change in the reward function of the consumers is seen to be lasting also in 2023, although in a less marked way compared to 2022, and with the registered empirical consumption also decreasing from 2022 levels. The change is therefore persistent rather than transient: this is seen also from the daily mean consumption, falling from 0.337 kW in 2021 to 0.329 kW in 2022 and 0.285 kW in 2023. \\
Moving to the urban intermediate-consumption cluster (Figure \ref{UrbanIntermediateConsAnalysis}, AC ownership 52.5\%): this group of consumers in 2021 has a more moderate response in temperature, clearly noticeable only above 30°C UTCI value from the reward's optimal consumption plot. This response, however, amplifies significantly in 2022 according to the reward function, while it is less obvious by looking at the empirical values trends. Then, it substantially reverts in 2023, where optimal consumption values fall back to 2021 trends on average, with slightly higher consumption for medium UTCI levels (between 22° and 30°) and slightly lower in the strong thermal stress region (> 30°). The change for this cluster is thus fast but transient, the 2022 amplification at high UTCI being completely lost the following year. \\
Finally, the low consumption clusters are considered. The rural low-consumption cluster (Figure \ref{RuralLowConsAnalysis}, AC ownership 16.1\% in 2021) stands out for its mostly flat, slightly decreasing, optimal consumption response to thermal stress in 2021: this behavioural phenomenon is not observable from empirical average data, where other factors, most likely the time of the day or other behavioural features, hide the signal. Also, for this cluster the onset of the energy crisis, along with higher thermal stress, lead to a change in both empirical and optimal computed consumption levels: first, an increase in 2022, in particular in the strong thermal stress region, above 30°; then, a consolidation of a more UTCI-responsive behaviour in 2023 with both empirical consumption and the reward signal showing decreases in consumption below 30° UTCI and increases above this threshold. The trajectory is therefore slow but structural, the moderate 2022 change being reinforced rather than reverted in 2023. \\
The urban low-consumption cluster (Figure \ref{UrbanLowConsAnalysis}, AC ownership 25.6\%) is the group of consumers with the most limited variation in response across the considered years. In 2021, it shows a weak but clear consumption response to thermal stress, both from the softmax-optimal and the empirical consumption. In 2022, the significant increase in thermal stress leads to a modest increase in optimal consumption above 30 °UTCI, which is then largely lost in 2023. In general, both the softmax-optimal and empirical consumption carry a noisier signal for this group of users, as can be seen from the large standard error of the mean empirical signal for several specific UTCI levels: this is the least responsive cluster of all, with small, directionless variations across the three years.

\subsection*{Impact of intraday variations on overall behaviour}
As mentioned in the results on clustering (Section \ref{sec:ResultsClustering}), the obtained clusters have well-defined context variable differences, which reflects the relative weight that this work's methodology has attributed to them (50\% variance, as stated in Section \ref{sec:Clustering}). So, in order to emphasize more the differences in consumption at sub-daily temporal resolution for specific groups of consumers, in this section the results on a sub-clustering of the high-consumption cluster are presented. To sub-cluster, the variance weighting has been modified, giving 50\% weight to electricity consumption and therefore reducing the importance of UTCI time series and of all the other context variables. \\
From the resulting 7 sub-clusters, two have been identified which present particularly interesting features at the sub-daily scale: sub-cluster 4 and sub-cluster 6. These two sub-clusters are almost indistinguishable for what concerns most of the considered variables (mean consumption, urbanization, GVI, income distribution, UTCI), but they do show a noticeable difference for estimated AC ownership and average intradaily trends of electricity consumption, as shown in Figure \ref{intradailyDifferences}. Sub-cluster 4 (\textbf{high-consumption, afternoon-high}) presents higher consumption during the day at all hours but especially in the afternoon, with a higher daytime response up to the [16--20] hours bin, while sub-cluster 6 (\textbf{high-consumption, evening-high}) presents a consumption profile skewed more towards evening consumption, with a more pronounced peak in the [20--24] hours bin. Also, the afternoon-high cluster reports a higher AC ownership, although both clusters are estimated to be heavy AC users. Following this, the two sub-clusters have been used as inputs to the AIRL pipeline, and the results are presented in the right panel of Figure \ref{intradailyDifferences} (as for the other clusters in the previous section). \\ As can be seen by the 2021 column and by the histogram in the bottom-left of the Figure, the clusters seem to experience very similar levels of thermal stress, and to have very similar empirical and optimal consumption levels in 2021. However, the situation changes significantly in 2022: the afternoon-high cluster, with more ACs, decreases its consumption at high thermal stress more strongly than the evening-high cluster with fewer ACs, being the group most willing to cut consumption at the 2022 crisis and thermal-stress peak; this second cluster instead decreases its consumption at lower thermal stress levels, but shows a mostly unchanged response at high UTCI, compared with 2021. Then, if 2023 results are considered, it can be seen that the afternoon-high cluster behaviour remains largely the one of 2022, while the evening-high cluster one flattens significantly, showing little thermal stress response in consumption: in other terms, they converge to a similar response in temperature.

\subsection{Discussion}\label{sec:Discussion}
The retrieved reward functions reveal distinct behavioural thermal responses, or cooling behaviours, across consumer groups. Relative to the gaps identified in the Introduction, this characterizes cooling behaviour at a granularity and with an interaction structure beyond the reach of qualitative or setpoint-based approaches \cite{congUnveilingHiddenEnergy2022}; and, unlike prior energy-crisis work based on stated practices \cite{chatzikonstantinouHousingEnergyConsumption2022}, it expresses the change during the crisis directly and quantitatively, as the consumption response to thermal stress.\\
The clusters differ both in the magnitude of their response and in its temporal dynamics. For the high-consumption cluster, the persistent decline and incomplete recovery of the strong 2021 response suggest that past high-consumption habits have been durably resized by the energy crisis, the full 2021 response never being recovered. The urban intermediate-consumption cluster changes instead fast but transiently: in a hot, low-GVI urban setting of old buildings, these users have almost no cooling option other than AC, and they have the financial means to use it; yet their weak 2021 baseline reasserts itself in 2023, as temperatures fall back toward 2021 levels.\\ The two low-consumption clusters diverge sharply despite a similarly weak response. The rural one follows a structural trajectory consistent with the large potential for cooling-related growth where AC is still little used, regardless of the crisis. The urban one is the least responsive of all, plausibly because the urban environment and crisis-tightened budgets leave little room for adaptation. That this urban low-consumption cluster and the urban intermediate one respond so differently in a similar environment (urban, high thermal stress, low GVI, old buildings) underlines how AC ownership and financial means mediate whether thermal stress translates into a consumption response.\\ Within the high-consumption cluster, the sub-clusters respond differently (Figure \ref{intradailyDifferences}). The afternoon-high one, with higher AC ownership and a higher daytime response, plausibly groups users more active at home during the day and thus cooling more \cite{stikvoortRhythmHomeHow2025a}, which would also explain why it is the group most willing to cut consumption at the 2022 peak. The evening-high one, more reluctant to change in the short term, retains more of its high-UTCI response in 2022 before converging to the afternoon-high behaviour in 2023. This may reflect time-of-use differences: in 2022, households more active in the afternoon can reschedule activities, postponing them to the evening or anticipating them in the morning, more easily than those whose activities already peak in the evening. Such intraday rescheduling under thermal stress is documented \cite{fanIntradayAdaptationExtreme2023}\cite{baturUnderstandingHowExtreme2024}, but its link to cooling and electricity consumption remains poorly characterized.\\ Finally, although this study cannot reconstruct the specific socioeconomic profiles of users within clusters, as done in more targeted energy-poverty work \cite{congUnveilingHiddenEnergy2022}\cite{kwonForgoneSummertimeComfort2023}\cite{huangInequalitiesCoolingHeating2023}, it does reveal broad differences in how cooling behaviours evolve under combined socioeconomic and thermal stress.
\subsection{Limitations and future work}
The current work is limited by the data availability and resolution of some of the used context variables. In particular, GVI, urbanization, UTCI, building vintages, income distributions cannot be considered at municipal or zip code granularity, which limits their explanatory power when single consumers are considered. Future work should consider data sources able to characterize each user from a socioeconomic and technical point of view, to overcome this limitation. A second limitation is methodological. Because the recovered reward is only partially identified, as discussed in the Methodology, the absolute level and scale of the reward does not necessarily carry an intrinsic meaning and should not be interpreted on their own. Our analysis therefore relies on comparisons across clusters and across years, under the defined constraints of the methodology, which remain robust to this indeterminacy even though the underlying reward is not point-identified. \\ From the methodological point of view, the current Adversarial Inverse Reinforcement Learning pipeline could benefit from the use of Long Short-Term Memory architectures for the PPO generator, to improve the reproduction of short-term (intraday) trends in electricity consumption. More methodological updates, such as a discretization of the training environment, could also improve the current model performance. \\
Then, a wider use of the reward function instrument, considering, for example, the study of higher-order partial derivatives, could be explored in order to deepen the study of single clusters. Finally, we acknowledge that billed energy cost and changes in contracts affecting consumers' consumption habits are factors that have been studied elsewhere with different methodologies \cite{lunghiPowerPlayBalancing2026} and that are not treated by this work. An analysis of how these factors interact with context variables and temperature should be considered in future studies on residential electricity consumption, making use of methodologies based on this study.

\section{Conclusion}\label{sec:Conclusion}
Throughout this work, the focus has been on understanding how households consume electricity considering socioeconomic and environmental drivers, and especially during summer periods when cooling demand can be a major driver of electricity consumption \cite{harishImpactTemperatureElectricity2020}.
To reach this, we have used Inverse Reinforcement Learning to represent household consumption behaviour as model-implied reward functions, from which optimal consumption curves have been obtained.\\ These curves present distinct shapes for the different groups of consumers that have been considered: clusters with high estimated air conditioning shares and higher-than-average consumption react strongly to temperature in 2021, and their consumption habits change significantly when going through the high thermal stress and socioeconomic shocks of 2022, during the peak of the energy crisis. How persistent this change is, when looking at 2023 data, depends on the specific cluster. \\
When considering intraday differences for high-consumption clusters, we find that groups of consumers with different consumption timings during the day (afternoon-high vs evening-high) change their consumption differently in the short term: in particular, users with an evening consumption peak tend to maintain a consumption behaviour strongly responding to thermal stress, while this is not true for users with an afternoon peak. This indicates that the timing of daily activity shapes how cooling demand behaviour responds to shocks.\\
This study contributes to the literature by showcasing a novel instrument based on Inverse Reinforcement Learning to study large datasets of smart meter data \cite{uchidaAggregatedSmartMeter2025}, that expands on the standard approaches currently used to study energy behaviour and energy poverty related to cooling demand \cite{kwonForgoneSummertimeComfort2023}. Unlike empirical consumption data, which reveals only the enactment of behaviour entangled with time-of-day and other confounding effects, this approach yields a reward representation of the temperature response that is disentangled from these factors. \\
This work also highlights the need for careful consideration by decision makers and designers of energy policies of how consumption behaviour might change through environmental and socioeconomic shocks \cite{wangIncentiveBasedEmergency2023} respond. If the consumer response function itself changes in the medium term because of thermal stress and external factors, this means that from one year to another, under the same environmental conditions, users might response differently in terms of electricity consumption. Therefore, demand-response and tax schemes need to account for this dynamic nature of users' behaviour. \\
Finally, this work emphasizes the relevance of the surrounding built environment and context in determining the consumption behaviour of households, and their evolution in time: from the results, we see that low-consumption users located in greener and rural areas tend to develop responses to thermal stress only starting in 2022 with periods of high exposure to high temperatures, while corresponding low-consumption households in urban environments exposed to similar shocks are not substantially changing their behaviour across the years. This not only motivates more in-depth research to study cooling behaviours and the local drivers of the uptake of cooling devices \cite{falchettaInequalitiesGlobalResidential2024}, but it can also enrich more long-term modeling approaches \cite{edelenboschTranslatingObservedHousehold2022}\cite{falchettaInequalitiesGlobalResidential2024}, where assumptions on the current and future consumers' behaviour is fundamental to properly design efficient energy and climate policies.

\section*{Author contributions}
Enrico Cofler: Conceptualization, Data curation, Formal analysis, Methodology, Software, Validation, Visualization, Writing - original draft \\
Carlos Rodriguez-Pardo: Conceptualization, Formal analysis, Supervision, Writing - review and editing \\
Matteo Giuliani: Conceptualization, Formal analysis, Supervision, Writing - review and editing \\
Andrea Castelletti: Conceptualization, Formal analysis, Supervision, Writing - review and editing \\
Massimo Tavoni: Conceptualization, Formal analysis, Supervision, Writing - review and editing, Project administration

\section*{Acknowledgments}
This study received support from the Energy Demand changes Induced by Technological and Social innovations (EDITS) project, an initiative coordinated by the Research Institute of Innovative Technology for the Earth (RITE) and International Institute for Applied Systems Analysis (IIASA) and funded by Ministry of Economy, Trade, and Industry (METI), Japan.

\section*{Declaration of competing interest}
The authors confirm that they do not have any conflicts of interest to declare.

\section*{Declaration of Generative AI and AI-assisted technologies in the writing process}
During the preparation of this work the author(s) used Grammarly in order to proofread the manuscript. After using this tool/service, the author(s) reviewed and edited the content as needed and take(s) full responsibility for the content of the publication.

\section*{Data availability}
The repository will be shared publicly once the paper is accepted.
Smart meter data, being private data, cannot be shared publicly.

\end{document}